\documentclass{article}
\usepackage{spconf,amsmath,graphicx,balance} 

\title{Skin Lesion Classification Using Hybrid Deep Neural Networks}

\name{
Amirreza Mahbod$^{\star \dagger}$\thanks{This research has received funding from the Marie Sklodowska-Curie Actions of the European Union's Horizon 2020 programme under REA grant agreement no. 675228.}, 
Gerald Schaefer$^{\ddagger}$,
Chunliang Wang$^{\mathsection}$, 
Rupert Ecker$^{\dagger}$, 
Isabella Ellinger$^{\star}$
}
\address{
$^{\star}$Institute for Pathophysiology and Allergy Research, Medical University of Vienna, Austria\\
$^{\dagger}$Department of Research and Development, TissueGnostics GmbH, Austria\\
$^{\ddagger}$Department of Computer Science, Loughborough University, U.K.\\
$^{\mathsection}$Department of Biomedical Engineering \& Health Systems, KTH Royal Institute of Technology, Sweden
}

\begin{document}

\maketitle

\begin{abstract}
Skin cancer is one of the major types of cancers with an increasing incidence over the past decades. Accurately diagnosing skin lesions to discriminate between benign and malignant skin lesions is crucial to ensure appropriate patient treatment. While there are many computerised methods for skin lesion classification, convolutional neural networks (CNNs) have been shown to be superior over classical methods. In this work, we propose a fully automatic computerised method for skin lesion classification which employs optimised deep features from a number of well-established CNNs and from different abstraction levels. We use three pre-trained deep models, namely AlexNet, VGG16 and ResNet-18, as deep feature generators. The extracted features then are used to train support vector machine classifiers. In a final stage, the classifier outputs are fused to obtain a classification. Evaluated on the 150 validation images from the ISIC 2017 classification challenge, the proposed method is shown to achieve very good classification performance, yielding an area under receiver operating characteristic curve of 83.83\% for melanoma classification and of 97.55\% for seborrheic keratosis classification.
\end{abstract}


\section{Introduction}
\label{sec:intro}
Skin cancer is one of the most common cancer types worldwide~\cite{Oliveira2018}. As an example, skin cancer is the most common cancer type in the United States and it is estimated that one in five Americans will develop skin cancer in their lifetime. Among different types of skin cancers, malignant melanoma (the deadliest type) is responsible for 10,000 deaths annually just in the United States~\cite{Rogers2015}. However, if detected early it can be cured through a simple excision while diagnosis at later stages is associated with a greater risk of death - the estimated 5-year survival rate is over 95\% for early stage diagnosis, but below 20\% for late stage detection~\cite{Esteva2017}.

There are a number of non-invasive tools that can assist dermatologists in diagnosis such as macroscopic images which are acquired by standard cameras or mobile phones~\cite{Oliveira2018}. However, these images usually suffer from poor quality and resolution. Significantly better image quality is provided by dermoscopic devices which have become an important non-invasive tool for detection of melanoma and other pigmented skin lesions. Dermoscopy supports better differentiation between different lesion types based on their appearance and morphological features~\cite{SBSWP93a}.

\begin{figure}
	\centering
	\includegraphics[height=4cm]{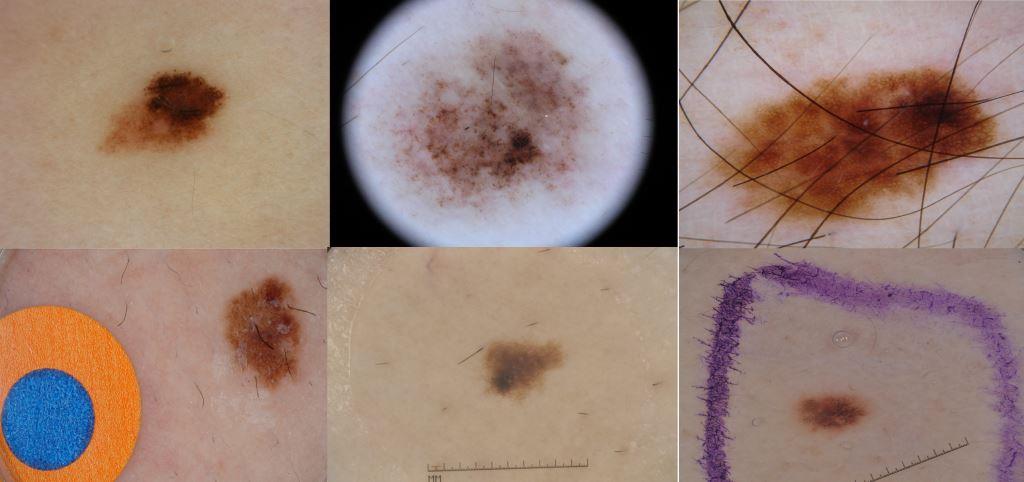}
	\caption{Common artefacts in dermoscopic images from the ISIC challenge. Normal image, dark corner artefact, skin hair artefacts, colour chart artefact, ruler marker artefact, ink marker artefact (left to right, top to bottom).}
	\label{artifact}
\end{figure}

Visual inspection of dermoscopic images is a challenging task that relies on a dermatologist's experience. Despite the definition of commonly employed diagnostic schemes such as the ABCD rule~\cite{SRCPAHBNLB94a} or the 7-point checklist~\cite{AFCDSD98a}, due to the difficulty and subjectivity of human interpretation as well as the variety of lesions and confounding factors encountered in practice (see Fig.~\ref{artifact} for some examples of common artefacts encountered in dermoscopic images), computerised analysis of dermoscopic images has become an important research area to support diagnosis~\cite{FSZGCPD98a}. Conventional computer-aided methods for dermoscopic lesion classification typically involve three main stages: segmenting the lesion area, extracting hand-crafted image features from the lesion and its border, and classification~\cite{CKUIASM07a}. In addition, often extensive pre-processing is involved to improve image contrast, perform white balancing, apply colour normalisation or calibration, or remove image artefacts such as hairs or bubbles~\cite{Oliveira2018, Barata2015}.

With the advent of deep convolutional neural networks (CNNs) and considering their excellent performance for natural image classification, there is a growing trend to utilise them for medical image analysis including skin lesion classification~\cite{Lopez2017}. Likewise, in this paper, we exploit the power of deep neural networks for skin lesion classification. Using CNNs, which are pre-trained on a large dataset of natural images, as optimised feature extractors for skin lesion images can potentially overcome the drawbacks of conventional approaches and can also deal with small task-specific training datasets. A number of works~\cite{Kawahara2016,Codella2015,Mirunalini2017,Lopez2017} have tried to extract deep features from skin lesion images and then train a classical classifier. However, these studies are limited by exploiting specific pre-trained network architectures or using specific layers for extracting deep features. Also, the utilised pre-trained networks were limited to a single network. In~\cite{Kawahara2016}, a single pre-trained AlexNet was used while~\cite{Lopez2017} employed a single pre-trained VGG16, and~\cite{Mirunalini2017} utilised a single pre-trained Inception-v3~\cite{szegedy2016rethinking} network.  

In this work, we hypothesise that using different pre-trained models, extracting features from different layers and ensemble learning can lead to classification performance competitive with specialised state-of-the art algorithms. In our approach, we utilise three deep models, namely AlexNet~\cite{Krizhevsky2012}, VGG16~\cite{Simonyan2014} and ResNet-18~\cite{He2016}, which are pre-trained on ImageNet~\cite{Deng2009}, as optimised feature extractors and support vector machines, trained using a subset of images from the ISIC archive\footnote{https://www.isic-archive.com/\#!/topWithHeader/wideContentTop/main}, as classifiers. In the final stage, we fuse the SVM outputs to achieve optimal discrimination between the three lesion classes (malignant melanoma, seborrheic keratosis and benign nevi).

\section{Materials and Methods}
\subsection{Dataset}
We use the training, validation and test images of the ISIC 2016 competition~\cite{gutman2016skin} as well as the training set of the ISIC 2017 competition\footnote{https://challenge.kitware.com/\#phase/5840f53ccad3a51cc66c8dab} for training the classifiers. In total, 2037 colour dermoscopic skin images are used which include 411 malignant melanoma (MM), 254 seborrheic keratosis (SK) and 1372 benign nevi (BN). The images are of various sizes (from $1022\times767$ to $6748\times4499$ pixels), photographic angles and lighting conditions and different artefacts such as the ones shown in Fig.~\ref{artifact}. A separate set of 150 skin images is provided as a validation set. It is these validation images that we use to evaluate the results of our proposed method. 

\subsection{Pre-processing}
\begin{figure*}
\centering
\includegraphics[width=\textwidth,trim={0 0 0 15cm},clip]{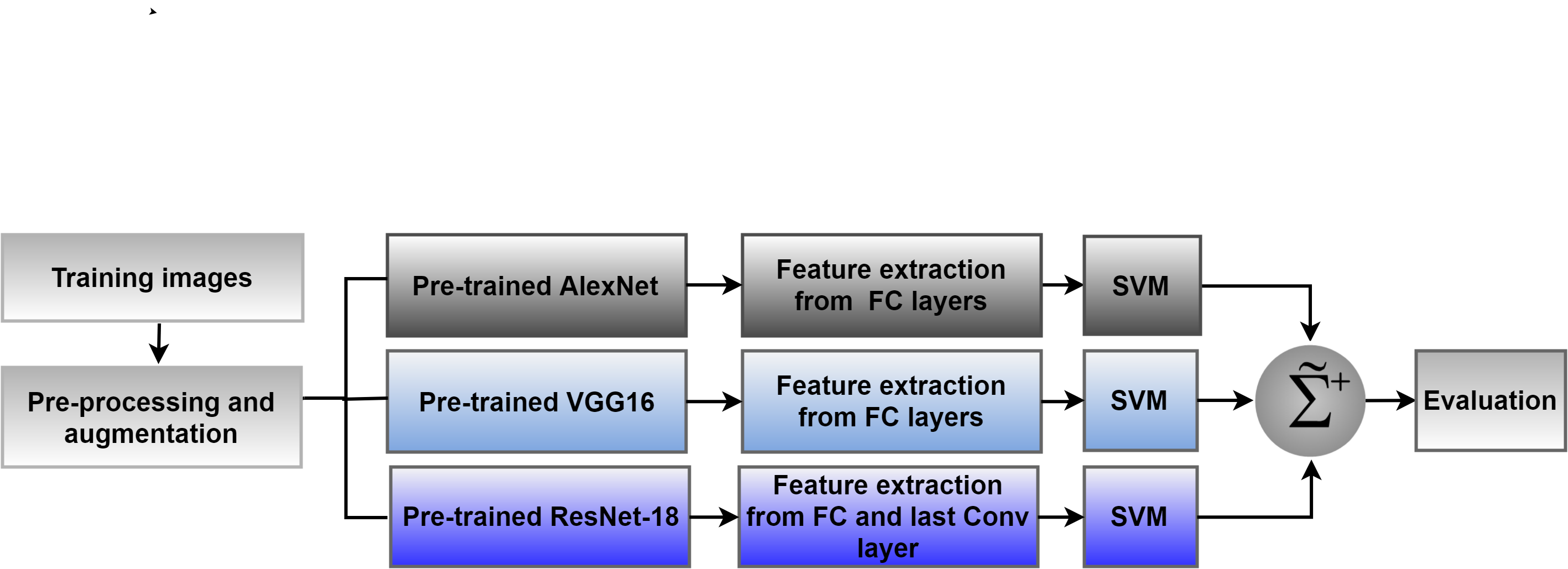}
\caption{Flowchart of the proposed method.}
\label{flowchart}
\end{figure*}

A generic flowchart of our proposed approach is shown in Fig.~\ref{flowchart}.

In our approach, we try to keep the pre-processing steps minimal to ensure better generalisation ability when tested on other dermoscopic skin lesion datasets. We thus only apply three standard pre-processing steps which are generally used for transfer learning. First, we normalise the images by subtracting the mean RGB value of the ImageNet dataset as suggested in~\cite{Krizhevsky2012} since the pre-trained networks were originally trained on those images. Next, the images are resized using bicubic interpolation to be fed to the networks ($227\times227$ and $224\times224$). Finally, we augment the training set by rotating the images by 0, 90, 180 and 270 degree and then further applying horizontal flipping. This augmentation leads to an increase of training data by a factor of eight.

\subsection{Deep Learning Models}
Our deep feature extractor uses three pre-trained networks. In particular, we use AlexNet~\cite{Krizhevsky2012}, a variation of \mbox{VGGNet} named VGG16~\cite{Simonyan2014}, and a variation of ResNet named ResNet-18~\cite{He2016} as optimised feature extractors. These models have shown excellent classification performance for natural image classification in the Image Large Scale Visual Recognition Challenges~\cite{Russakovsky2015} and various other tasks. We choose the shallowest variations of VGGNet and ResNet to prevent overfitting since the number of training images in our study is limited. We explore extracting features from different layers of the pre-trained models to see how they can affect classification results. The features are mainly extracted from the last fully connected (FC) layers of the pre-trained AlexNet and pre-trained VGG16. We use the first and second fully connected layers (referred to as FC6 and FC7 with dimensionality 4096) and the concept detector layer (referred to as FC8 with dimensionality 1000). For ResNet-18, since it has only one FC layer, we also extract features from the last convolutional layer of the pre-trained model. 

\subsection{Classification and Fusion}
The above features along with the corresponding labels (i.e., skin lesion type) are then used to train multi-class non-linear support vector machine (SVM) classifiers. We train different classifiers for each network and then, to fuse the results, average the class scores to obtain the final classification result. To evaluate the classification results,  we map SVM scores to probabilities using logistic regression~\cite{Platt1999}. Since the classifiers are trained for a multi-class problem with three classes, we combine the scores to yield results for the two binary classification problems defined in the ISIC 2017 challenge, which are malignant melanoma vs.\ all and seborrheic keratosis vs.\ all classifications.

\section{Results}
As mentioned above, evaluation is performed based on the 150 validation images provided by the ISIC 2017 challenge. The validation set comprises 30 malignant melanoma, 42 seborrheic keratosis and 78 benign nevus images. For evaluation, we employ the suggested performance measure of area under the receiver operating characteristics curve (AUC). The raw images are resized to $227\times227$ pixels for AlexNet and to $224\times224$ pixels for VGG16 and ResNet-18. For each individual network and also for each fusion scheme, the results are derived by taking the average of the outputs over 5 iterations.


The obtained classification results are shown in Table~\ref{result} for all single networks and for all fused models.

\begin{table*}
\caption{Experimental results on ISIC 2017 validation dataset.}
\centering
\label{result}
\begin{tabular}{llccc}
\hline
\bf network & \bf feature layers & \bf MM AUC & \bf SK AUC  & \bf avg. AUC \\
\hline
AlexNet          &FC8     & 80.67   & 94.95   & 87.81  \\
AlexNet          &all FC  & 82.81   & 96.65    & 89.73 \\
VGG16          &FC8     & 82.61  & 90.94   & 86.78  \\
VGG16          &all FC  & 82.06 & 95.46  & 88.76  \\
ResNet-18          &FC & 81.00   & 91.93  & 86.47 \\
ResNet-18          &FC + last convol. layer    & 82.81  & 94.22 & 88.51   \\
\hline		
AlexNet + VGG16 fusion    &all FC   & 83.56    & 97.05   & 90.30   \\
AlexNet + ResNet-18 fusion    &all FC   & 83.53   & 97.05   & 90.29  \\
VGG16 + ResNet-18 fusion    &all FC   & 83.69   & 95.97   & 89.83   \\
\hline
fusion of all networks    &all FC   & \bf 83.83    &\bf 97.55 & \bf 90.69   \\
\hline
\end{tabular}
\end{table*}

\begin{figure}
\centering
\includegraphics[width=\columnwidth]{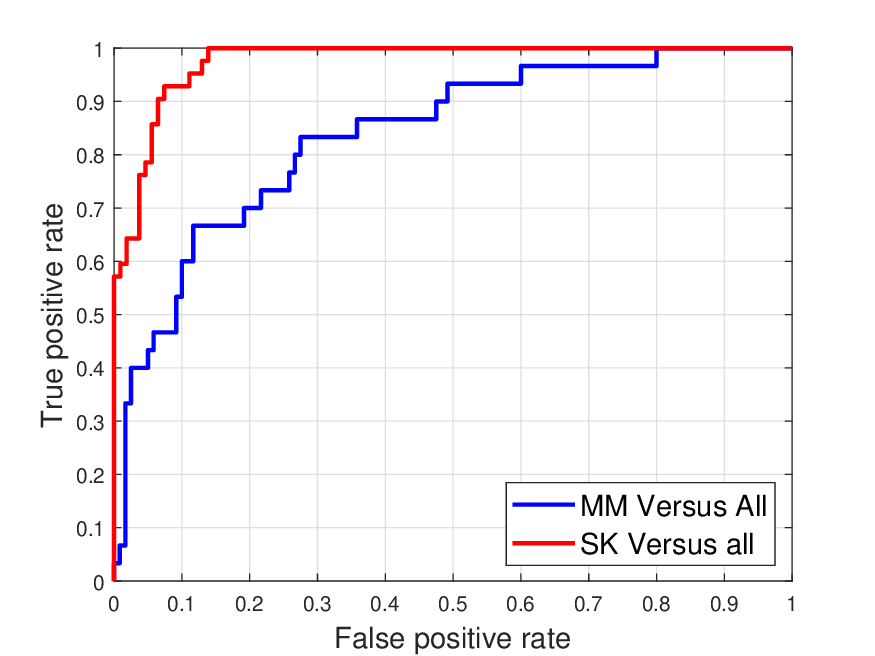}
\caption{ROC curve of the best performing approach.}
\label{ROC}
\end{figure}
\begin{figure}
\centering
\includegraphics[width=\columnwidth]{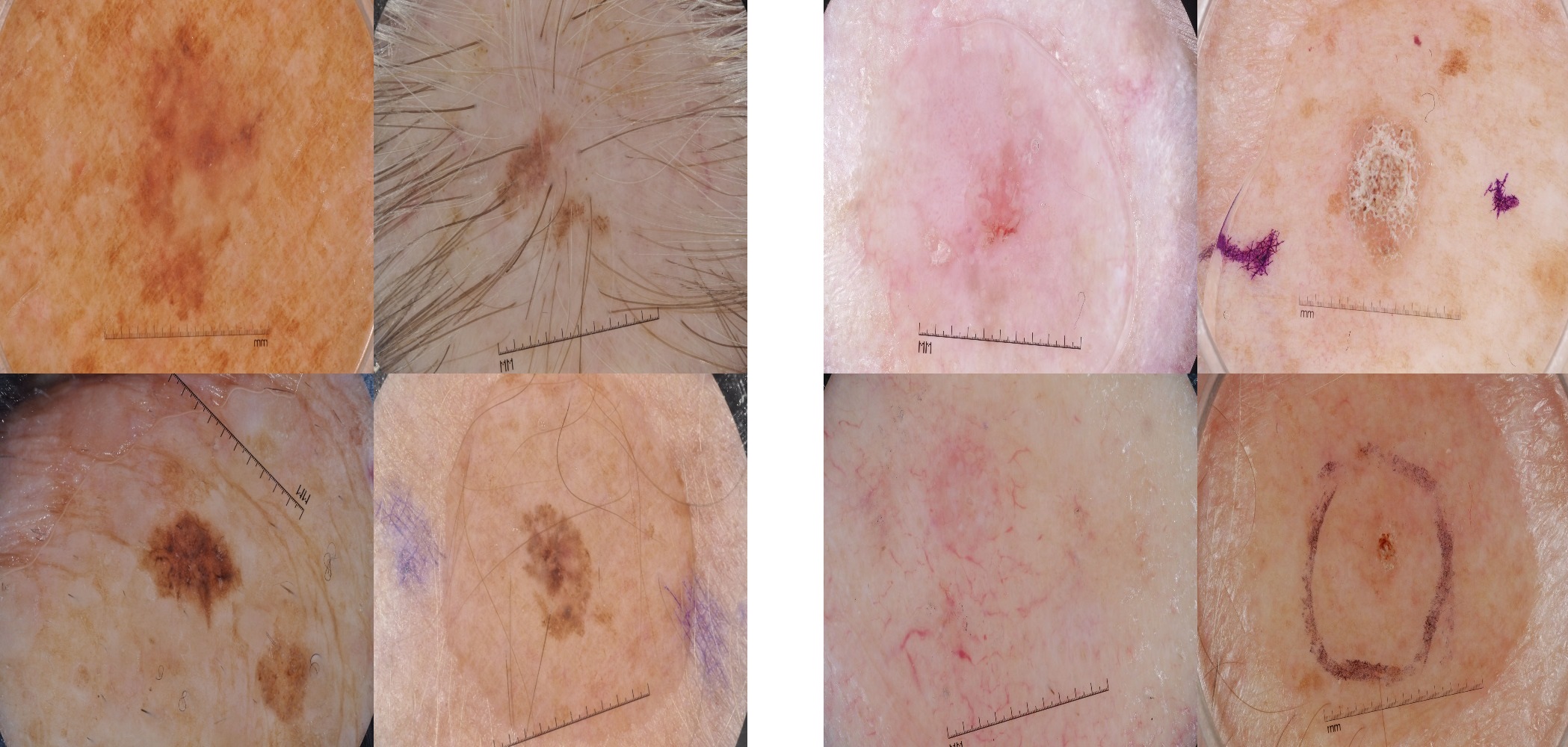}
\caption{Examples of incorrectly classified images for malignant melanoma vs.\ all (left) and seborrheic keratosis vs.\ all (right) tasks. }
\label{FN}
\end{figure}

Fig.~\ref{ROC} shows the receiver operating characteristic (ROC) curve of our best performing approach (i.e., fusion of all networks) while Fig.~\ref{FN} show examples of skin lesion images that are incorrectly classified by this approach. 

\section{Discussion}
The main contribution of this study is proposing a hybrid approach for skin lesion classification based on deep feature fusion, training multiple SVM classifiers and combining the probabilities for fusion in order to achieve high classification performance. 

From the classification results in Table~\ref{result}, we can infer a number of observations. First of all, for all approaches, even for the worst performing approach, the classification results are far better than pure chance (i.e. AUC of 50\%) which confirms that the concept of transfer learning can be successfully applied to skin lesion classification. Besides this, for all single networks, fusing the features from different abstraction levels leads to better classification performance compared to extracting features from a single FC layer.

Features extracted from AlexNet lead to the best performance of a single network approach. This could be potentially related to the network depth. Since our training dataset is not very big, using a shallower network may lead to better results.

The single network approaches are however outclassed by our proposed method of employing multiple CNNs and fusing their SVM classification outputs. The obtained results demonstrate that significantly better classification performance can be achieved. 



While our proposed method is shown to give very good performance on what is one of the most challenging public skin lesion dataset, there are some limitations that can be addressed in future work. First, the number of pre-trained networks that we have studied so far is limited. Extending the model to incorporate more advanced pre-trained models such as DenseNets~\cite{huang2017densely} could lead to further improved classification performance. Second, extending the training data is expected to lead to better results for each individual network as well as their combinations. Finally, resizing the images to very small patches might removing some useful information from the lesions. Although in a number of studies bigger training patches were used (e.g.\ $339\times339$ in~\cite{Kawahara2016} or $448\times448$ in~\cite{DeVries2017}), these are still significantly smaller compared to the captured image sizes. Cropping the images or using segmentation masks to guide the resizing could be a potential solution for dealing with this.    

\section{Conclusions}
In this paper, we have proposed a fully automatic method for skin lesion classification. In particular, we have demonstrated that pre-trained deep learning models, trained for natural image classification, can also be exploited for dermoscopic image classification. Moreover, fusing the deep features from various layers of a single network or from various pre-trained CNNs is shown to lead to better classification performance. Overall, very good classification results have been demonstrated on the challenging images of the ISIC 2017 competition, while in future work fusing more deep features also from further CNNs can potentially lead to even better predictive models.

\balance
\bibliographystyle{IEEEbib-abbrev}

\end{document}